\documentclass[preprint,10pt]{elsarticle}

\usepackage{amssymb}
\usepackage{amsmath}
\usepackage{booktabs}
\usepackage{multirow}

\journal{Data \& Knowledge Engineering}

\usepackage{lineno}
\begin{document}

\begin{frontmatter}


\title{SASA: Semantic-Aware Contrastive Learning Framework with Separated Attention for Triple Classification}
\author{Xiaodan Xu}
\ead{diane1968@163.com}
\author{Xiaolin Hu\cormark[cor1]}
\ead{huxiaolin0720@163.com}
\cortext[cor1]{Corresponding author}
\affiliation{organization={Aerospace Science \& Industry Defense Technology Research and Test Center},
            addressline={Yongding Road},
            city={Beijing},
            postcode={100039},
            country={China}}




\begin{abstract}
Knowledge Graphs~(KGs) often suffer from unreliable knowledge, which restricts their utility. Triple Classification~(TC) aims to determine the validity of triples from KGs. Recently, text-based methods learn entity and relation representations from natural language descriptions, significantly improving the generalization capabilities of TC models and setting new benchmarks in performance. However, there are still two critical challenges. First, existing methods often ignore the effective semantic interaction among different KG components. Second, most approaches adopt single binary classification training objective, leading to insufficient semantic representation learning.
To address these challenges, we propose \textbf{SASA}, a novel framework designed to enhance TC models
via separated attention mechanism and semantic-aware contrastive learning~(CL). Specifically, we first propose separated attention mechanism to encode triples into decoupled contextual representations and then fuse them through a more effective interactive way. Then, we introduce semantic-aware hierarchical CL as auxiliary training objective to guide models in improving their discriminative capabilities and achieving sufficient semantic learning, considering both local level and global level CL. Experimental results across two benchmark datasets demonstrate that SASA significantly outperforms state-of-the-art methods. In terms of accuracy, we advance the state-of-the-art by +5.9\% on FB15k-237 and +3.4\% on YAGO3-10.
\end{abstract}




\end{frontmatter}



\section{Introduction}
\label{sec_intro}
Knowledge graphs (KGs), such as Freebase \cite{freebase} and Wikidata \cite{wikidata}, organize structured facts that have become crucial resources of knowledge for various fields, such as recommendation systems \cite{surveykgrecommendersystems}, question answering \cite{interactivekgqa}, and information retrieval \cite{entityranking}. In general, a KG is composed of a collection of triples, with each triple represented as $(h, r, t)$ to encapsulate the specific relationship $r$ between two distinct entities, namely the head entity $h$ and the tail entity $t$. Factual knowledge is inherently infinite and perpetually evolving, which gives rise to concerns about the unreliability and incompleteness of KGs.

To tackle this issue, researchers have focused on Knowledge Graph Completion (KGC) models that focus on mitigating these inherent limitations of KGs. In this paper, we particularly target Triple Classification~(TC) for KGC, whose goal is to assess the validity of a given triple.

Many studies have been conducted in the field of TC, and these works fall into two main categories: embedding-based methods and text-based methods. Embedding-based
methods model connection patterns between entities in KGs, followed by scoring functions to determine the validity of triples \cite{transE,transH,rotate}. While pretrained language models (PLMs) have excelled in many natural language processing (NLP) tasks \cite{bert}, the limitations of embedding-based methods have become increasingly pronounced: these methods neither incorporate side information such as entity descriptions nor take contextualized information into account, thus resulting in a bottleneck in performance improvement. In contrast, text-based methods collect valid texts for entities and relations, and leverage preliminary linguistic information from PLMs to boost performance \cite{kgrepresentation,kepler,kgbert}.

Despite their effectiveness, current text-based studies on TC still face two critical challenges. First, existing methods
often ignore the effective semantic interaction among the representations of head entities, tail entities and relations,
where these three components typically interact with each other via models' native token-level cross-attention. This interaction paradigm severely limits the effective information flow between distinct representation subspaces. Second, recent approaches have applied contrastive learning~(CL) for link prediction KGC models, significantly improving their discriminative power; however, limited studies have explored the integration of contrastive approaches for the TC task, leading to insufficient semantic representation learning. What's more, current contrastive approaches usually perform instance discrimination in or between batches, which is not explicitly designed to identify the fine-grained semantics of knowledge in KGs, struggling to distinguish between entities with lexical similarity. This gap highlights the need for a new holistic paradigm to develop more robust TC models.

In this study, we propose a novel framework, named SASA, designed for Semantic-Aware contrastive triple classification with Separated Attention mechanism. Specially, SASA first uses a dual-tower architecture augmented with separate attention mechanism to enhance the semantic interaction among the decoupled representation spaces in a more flexible manner. Subsequently, a semantic-aware contrastive learning module is proposed to guide the model to effortlessly distinguish between fine-grained semantic differences in KGs. We introduce two contrastive learning tasks from a hierarchical perspective, considering both local-level and global-level modeling. During the finetuning phase of PLMs, we jointly train these tasks, which facilitates a collaborative reinforcement effect among different tasks and empowers the model to effectively capture the underlying fine-grained semantic information within KGs.

In summary, our contributions are as follows:
\begin{enumerate}
\item We propose a novel separated attention mechanism that offers the flexibility to enhance the semantic interaction among the decoupled representation spaces of KG components.

\item We creatively design a hierarchical contrastive learning module integrating four semantic-aware tasks, which achieves comprehensive semantic awareness at both local and global levels.

\item We conduct evaluations on two benchmark datasets. Experimental results and analyses demonstrate that our approach outperforms state-of-the-art methods in terms of effectiveness under standard settings.
\end{enumerate}
The rest of this paper is structured as follows: Section~\ref{related_work} reviews related works, followed by the methodology proposed in Section~\ref{methodology}. Section~\ref{experiment} presents the experimental results, and Section~\ref{conclusion} concludes with discussions on key findings and future research.


\section{Related works}
\label{related_work}
In recent years, the surge of interest in triple classification, a pivotal sub-task of KGC, has underscored its indispensable role in advancing knowledge graph-related research. As a fundamental module for refining knowledge graph quality, triple classification serves as a critical gateway for validating the authenticity of relational facts, mitigating noise accumulation, and laying a solid foundation for downstream tasks like link prediction and knowledge reasoning \cite{kgreasoning}. 

Consequently, researchers have devoted substantial efforts to developing robust triple classification models including embedding-based methods and text-enhanced approaches. 


\subsection{Embedding-based Methods}
Triple classification aims to predict whether a triple is valid or not. Embedding-based methods map entities and relations into a continuous vector space. Early approaches in this category typically rely on manually designed scoring functions to assess the validity of triples. TransE \cite{transE}, a seminal work in this domain, proposes the translation-based hypothesis, formulating each relation as a vector translation from the head entity to the tail entity in a low-dimensional space. Building on this intuition, RotatE \cite{rotate} extends the translation mechanism to the complex vector space, enabling the natural modeling of diverse relational patterns such as symmetry, inversion, and composition.

Beyond translation-based models, semantic matching methods prioritize capturing intricate interactions between entities and relations through vector-wise interactions. Representative works include DistMult \cite{distmult}, which employs bilinear diagonal matrices to model relational semantics, and ComplEx \cite{complex}, which leverages complex-valued embeddings to effectively handle asymmetric and inverse relations by incorporating both real and imaginary components. However, these methods treat each triple in isolation, overlooking the graph structure information inherent in KGs.

To further exploit the graph-structured nature of KGs, a subsequent line of research integrates graph neural networks (GNNs) to aggregate contextual structural information. RGCN \cite{rgcn} pioneers this direction by adapting graph convolutional networks (GCNs) for relational data, using relation-specific weight matrices to model neighborhood aggregation. SMiLE \cite{smile} proposed schema-aware attention mechanism to prioritize meaningful neighbor information and mitigate noise for KGC. Nevertheless, these models still have shortcomings: they overlook the potential semantic correlations among the knowledge graph contexts.

\subsection{Text-based Methods}
Text-based methods leverage descriptive information to capture the semantics of knowledge graph components. Early representative works include KG-BERT, StAR, SimKGC, and StructKGC \cite{kgbert,stAR,simkgc,structkgc}. Specifically, KG-BERT \cite{kgbert} fine-tunes BERT with cross-entropy loss to learn discriminative entity embeddings for KGC. StAR \cite{stAR} adopts a decoupled dual-encoder architecture and fine-tunes PLMs for efficient KGC. To capture fine-grained semantic correlations, SimKGC \cite{simkgc} and StructKGC \cite{structkgc} reformulate the link prediction task as a semantic matching problem and incorporate contrastive learning into the training process. But these methods are plagued by the issue of unstable negative sampling, which restricts their performance gains.

In the realm of generative KGC, researchers have achieved further advancements by leveraging sequence-to-sequence PLMs. Specifically, KG-S2S \cite{kgs2s} and UniLP \cite{unilp} utilize the T5 model \cite{t5} combined with soft prompt techniques to improve the quality of missing triple generation. CSProm-KG \cite{cspromkg} adopts a hybrid approach by integrating PLMs with traditional structure-based KGC methods, effectively bridging the semantic gap between textual information and graph structural knowledge. However, these methods primarily focus on the link prediction task rather than triple classification; the core goal of the former is to infer missing components (head entity, relation, or tail entity) of incomplete triples.

Moreover, LLMs have revolutionized numerous NLP tasks with their exceptional semantic understanding and knowledge reasoning capabilities, prompting increasing exploration of their potential in KGC. Early pioneering LLM-based works include KG-LLM \cite{kgllm}, who took a critical step by adopting instruction tuning to adapt LLaMA \cite{llama} and ChatGLM \cite{chatglm} specifically for KGC tasks. Recently, \cite{distillationkgc} proposed knowledge distillation from LLMs to refine the quality of entity textual descriptions, indirectly boosting the performance of existing text-based KGC methods. KoPA \cite{kopa} introduced a knowledge prefix adapter to project pretrained KG structural embeddings into LLM’s text token space, which enables structure-aware reasoning by incorporating subgraph information that was overlooked previously.

Despite the aforementioned success, LLM-based methods typically impose high computational demand due to their large parameter scales and complex training paradigms. Therefore, our work focuses on developing efficient small-model solutions tailored for triple classification task, aiming to balance model performance and computational feasibility.

\section{Methodology}
\label{methodology}
\begin{figure}
    \centering
    \includegraphics[width=\textwidth]{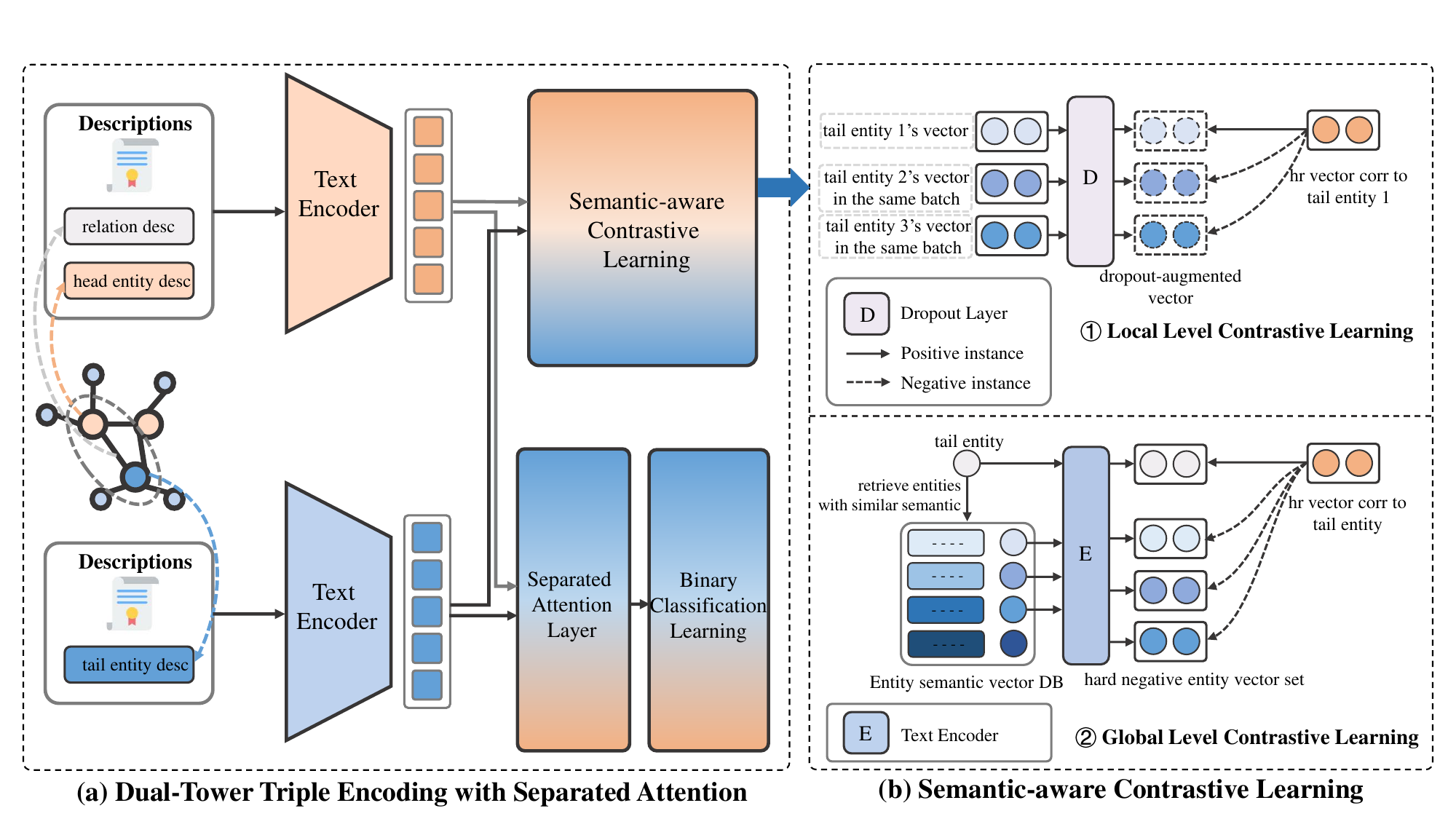}
    \caption{The overall framework of our \textbf{SASA}.} \label{fig_overview}
\end{figure}

Our study introduces a model, called Semantic-Aware contrastive triple classification with Separated Attention mechanism~(SASA), which is designed to enhance the accuracy and robustness of triple classification. This method involves two modules:

\textbf{Dual-Tower Triple Encoding with Separated Attention}: We encode the decoupled triples into separated contextual representations using a BERT based dual-tower model. Then we align heterogeneous representations into a unified semantic space via separated attention mechanism. The fused representation is quantified into scores that represent the likelihood of a triple being trustworthy.

\textbf{Semantic-Aware Contrastive Learning}: We combine binary classification training objective with CL for more sufficient semantic learning. From a hierarchical perspective, we design two complementary CL tasks to capture both local and global semantic patterns. At the local level, we feed the tail entity embedding into a dropout layer to generate noise-augmented embeddings with minimal semantic perturbation. We then propose a local-level CL task to align the head-relation pair $(h, r)$ with its dropout-augmented positive tail entity $t$ samples. At the global level, we first employ large embedding model BGE to retrieve hard negative entities for the anchor tail entity. To leverage the valuable insights offered by hard negative examples with high semantic similarity, We formulate the global-level CL task as teaching the model to discriminate the anchor entity from hard negatives. 

These modules are jointly trained to effectively capture the underlying fine-grained semantics within KGs.



\subsection{Task Formulation}
\label{task_formulation}
A knowledge graph $G$ is defined as a directed graph where vertices correspond to entities $E$, and each edge can be represented as a relation. Herein, $h$, $r$, and $t$ denote the head entity, relation, and tail entity, respectively. Moreover, entities are typically accompanied by rich textual descriptive information. In this paper, we focus on the TC task, whose core objective is to determine the validity of a given triple $(h, r, t)$\textemdash specifically, to judge whether the inter-entity relationship described by the triple truly exists in the KG.

\subsection{Dual-Tower Triple Encoding with Separated Attention}
\label{dual_tower_encoder}
\subsubsection{Dual-Tower Triple Encoding}
Our proposed model SASA adopts a bi-encoder architecture. Given a triple $(h,r,t)$, we first partition the triple into two components: one with a combination of head and relation $(h, r)$, and the other with tail $t$. Then, inspired by StAR~\citep{stAR}, a Siamese-style textual encoder is applied to the textual descriptions of these two components, encoding each into a distinct contextualized representation. 

Specifically, the first encoder, denoted as $BERT_{hr}$, is dedicated to generating the relation-aware embedding for the head entity. To this end, We concatenate the textual descriptions of the head entity $h$ and relation $r$, with a special separator token $[SEP]$ inserted between them. The concatenated sequence is formulated as follows:
\begin{equation}
\hat{x}_{hr} = \{x_{[CLS]}, x_h, x_{[SEP]}, x_r, x_{[SEP]}\}
\end{equation}
where $x_{[CLS]}$ and $x_{[SEP]}$ are special tokens and we use the segment identifier to distinguish whether a token originates from an entity (labeled as 0) or a relation (labeled as 1).

$BERT_{hr}$ is then fed the concatenated sequence to generate the last-layer hidden states $e^{i}_{hr} \in R^{d}$, where $i$ indexes the tokens in the input sequence. These hidden states collectively form the token-level representation set:
\begin{equation}
E^{hr} = \{e^{1}_{hr}, e^{2}_{hr}, ..., e^{L}_{hr}\}
\end{equation}
Herein, $L$ is the total length of the concatenated textual descriptions of the head entity $h$ and relation $r$. Also, $E^{hr}$ can be regarded as the contextualized token-wise representation that encodes the semantic information across the head entity and relation.

Similarly, the second encoder, denoted as $BERT_t$, is used to compute the $L_2$-normalized embedding $e_t$ for the tail entity $t$. The input to $BERT_t$ solely comprises the textual description of entity $t$, which is written as:
\begin{equation}
\hat{x}_{t} = \{x_{[CLS]}, x_{t}, x_{[SEP]}\}
\end{equation}

Rather than directly adopting the hidden state of the $[CLS]$ token, we employ mean pooling followed by $L_2$ normalization to get the tail embedding $e_t$. This design choice is motivated by empirical evidence that mean pooling yields superior sentence-level embeddings compared to $[CLS]$ token pooling \cite{sentencebert,simcse}. This process is formally defined as:
\begin{equation}
e_{t} = L_2(MeanPool(BERT_{t}(\hat{x}_{t})))\\
\end{equation}
where $MeanPool(\cdot)$ computes the mean of the last-layer hidden states to derive the sequence-level contextualized representation, and $L_2(\cdot)$ scales the mean pooling vector into a length of 1 via dividing each element by the vector's original $L_2$ norm.

Accordingly, $e_{t}$, the contextualized representation of the tail entity $t$, is viewed as tail embedding. We also get the head-relation embedding $e_{hr}$ in the same way. In our experiments, we keep the two BERT encoders parameter-separated for diverse representation learning. Notably, the encoder can be initialized with alternative PLMs to further enhance its representation learning capability\textemdash such as Transformer-XL \cite{transformerxl}, RoBERTa \cite{roberta}, or other state-of-the-art models. For illustrative purposes, we use BERT as the base encoder in our experiments.

\subsubsection{Separated Attention Mechanism}
In this section, we introduce a separated attention mechanism guided by a flexible cross-attention module to learn contextualized representation of a triple.

First, we align the decoupled representations (i.e., $e^{i}_{hr}$ and $e_{t}$) by projecting them into the same semantic space, enabling more effective interaction between the two components. The projected representations are computed as follows:
\begin{eqnarray}
h^{i}_{hr} &=& W_{1}e^{i}_{hr}+b_{1}\\
\widehat{h}^{i}_{hr} &=& W_{2}(Dropout(LayerNorm(GELU(h^{i}_{hr}))))+b_{2}\\
\widehat{H}_{hr} &=& \{\widehat{h}^{1}_{hr}, \widehat{h}^{2}_{hr}, \dots, \widehat{h}^{L}_{hr}\}\\
h_{t} &=& W_{1}e_{t}+b_{1}\\
\widehat{h}_{t} &=& W_{2}(Dropout(LayerNorm(GELU(h_{t}))))+b_{2}
\end{eqnarray}
where $W_{1} \in R^{d \times 2d}$, $b_{1} \in R^{2d}$, $W_{2} \in R^{2d \times d}$, $b_{2} \in R^{d}$ are trainable parameters for projections. In order to achieve a stable training process, we also employ GELU followed by layer normalization between two linear layers.

Given the above projected representations, we employ a cross-attention module with $\widehat{h}_{t}$ as the query to get the fused representation $h_{hrt}$, enabling the tail embedding to focus on semantically related parts of $\widehat{H}_{hr}$ and eliminate noisy aggregations:

\begin{eqnarray}
A &=& \frac{(\widehat{h}_{t}W^{Q})(\widehat{H}_{hr}W^K)^{T}}{\sqrt{d}} \\
h_{hrt} &=& Softmax(A)(\widehat{H}_{hr}W^{V})
\end{eqnarray}
where $d$ is the dimension of the query and key vectors, and $W^K$, $W^V$, $W^Q$ are trainable parameters.

\subsubsection{Triple Classification Training Objective}
Using the fused representation $h_{hrt}$, we further employ binary cross-entropy loss as the objective function to quantify the differences between the predicted confidence and the ground truth label as delineated below.

Log loss is particularly suitable for the binary classification task because it penalizes incorrect predictions logarithmically, encouraging the model to make more confident predictions. This loss function enables us to align the predicted confidences of triples (which range between 0 and 1) with the corresponding ground-truth labels (0 for invalid triples, 1 for valid ones). Let $W_{TC} \in R^{2 \times d}$ be a linear layer for triple classification, then:

\begin{eqnarray}
f(x) &=& Softmax(h_{hrt}W_{TC}^{T})=[\hat{y}_0, \hat{y}_1]\\
\mathcal{L}_{TC} &=& -\sum_{x \in \mathcal{D}} ylog\hat{y}_{1}+(1-y)log\hat{y}_{0}
\end{eqnarray}
where $f(x)$ is the final output distribution of the model for the input triple $x=(h,r,t)$; $y \in \{0, 1\}$ is the ground-truth label of the input $x$; $\hat{y}_1$ represents the predicted probability of the valid class, and $\hat{y}_1$ denotes the predicted probability of the invalid class.

\subsection{Semantic-Aware Contrastive Learning}
\label{cl}
\subsubsection{Revisiting Contrastive Learning}
Contrastive learning aims to learn effective representations by pulling semantically close neighbors together and pushing apart non-neighbors. This approach has been proven successful in the task of link prediction. To take additional contrastive learning supervision into considerations, we follow the contrastive learning framework in StructKGC \cite{structkgc}, generalizing the InfoNCE loss \cite{infonce} to support multi-positive contrastive learning:

\begin{equation}
    \mathcal{L}_{CL}=-\frac{1}{\lvert P(t) \rvert}\sum_{v \in P(t)}log \frac{e^{\phi(hr,v)/\tau}}{\sum_{i=1}^{\lvert \mathcal{N}\rvert}e^{\phi(hr,t_{i})/\tau}}
\label{eq1}
\end{equation}
where $\mathcal{N}$ represents a set of negative examples in the same batch and $P(t)$ consists of multiple positive tail samples semantically related to $hr$. The temperature parameter $\tau$ is introduced to control the relative significance of these negatives. Smaller $\tau$ makes the loss put more emphasis on hard negatives, but also risks over-fitting label noise. To avoid tuning $\tau$ as a hyperparameter, we re-parameterize $log\frac{1}{\tau}$ as a learnable parameter. What's more, cosine similarity is used as the scoring function to measure the distance between the two components:
\begin{equation}
    \phi(hr, t) = cos(e_{hr}, e_{t}) \in [-1, 1]
\end{equation}
where $hr$ denotes the query (i.e., an head-relation pair); $t$ is the corresponding tail entity of $hr$; $e_{hr}$ and $e_{t}$ denotes the encoded representations of $hr$ and $t$.

To guide the model to effectively capture fine-grained semantic differences within KGs and distinguish between candidates with lexical similarity, we integrate two semantic-aware contrastive learning tasks into the finetuning paradigm of PLMs, as illustrated in Fig.\ref{fig_overview}. Next, we will introduce these tasks and the detailed definitions of different types of semantic positive and negative samples in the following sections.

\subsubsection{Local-level CL}
A critical challenge in CL is how to construct positives. Given a query $(h,r)$, previous works \cite{simkgc,structkgc} usually adopt either the original corresponding tail entity $t$ or contexts surrounding $(h, r)$ as positives. However, semantic representations of nodes in KGs trained with only $t$ as positives are not robust, while utilizing contexts surrounding $(h, r)$ as positives may mislead the model, making it struggle to distinguish between candidates with lexical similarity. 

Inspired by SimCSE \cite{simcse}, we propose a local-level contrastive learning strategy to enable the model to learn more robust fine-grained semantic representations. Specifically, this strategy processes the tail entity $t$ with only dropout used as noise. Concretely, we feed the tail embedding $e_t$ into a standard dropout layer to obtain a dropout-augmented tail embedding $e'_t$ and treat it as positives for the query $(h,r)$. We regard other samples within the same batch as negatives, and the model identifies the positive sample among these negative samples. We hypothesize that dropout acts as a minimal form of semantic perturbation to the hidden representations, which can enhance the model's capacity to robustly capture the fine-grained semantics of node representations.

The key ingredient to get CL to work with noisy positives is through the use of the dropout mask for $e_t$. We add an additional dropout layer on top of $BERT_t$. Given a target entity $t$, we formally define its dropout-augmented positives as the set of dropout representations: 

\begin{equation}
    P_{local}(t) = \{e'_t|e'_t = f_{\theta}(e_t,z)\}
\end{equation}
where $z$ is a random dropout mask corresponding to the additional dropout layer atop the encoder $BERT_t$. Specifically, we feed the tail embedding $e_t$ to the dropout layer and get noisy embedding $e'_{t}$ as positives. Based on this, a local-level CL task is proposed to align the head-relation pair with its dropout-augmented positives. Following \eqref{eq1}, the loss of local-level CL is defined as:

\begin{equation}
    \mathcal{L}_{local}=-\frac{1}{\lvert P_{local}(t) \rvert}\sum_{e'_{t} \in P_{local}(t)}log \frac{e^{\phi(hr,e'_{t})/\tau}}{\sum_{i=1}^{\lvert \mathcal{N}\rvert}e^{\phi(hr,t_{i})/\tau}}
\end{equation}
for a batch of $\mathcal{N}$ samples.

In this way, We view the dropout mask as a form of minimal semantic perturbation: compared to original $e_t$, the dropout-augmented representation $e'_t$ is derived from the identical textual description of the tail entity $t$, with the only difference between the two embeddings lying in the applied dropout mask.

\subsubsection{Global-level CL}
Previous studies on CL usually take other samples in the same batch as negatives. However, these approaches tend to neglect the rich insights offered by hard negative examples with high semantic similarity. We observe that forcing the model to distinguish between hard negative pairs of extreme semantic similarity substantially improves its discrimination ability, which is frequently overlooked by the existing literature. Thus, to capture the fine-grained semantic differences conveyed by hard negative examples, we introduce a global-level CL strategy that enables the model to discriminate between the anchor entity and hard negatives. This two-stage process includes Hard Negative Pairs Construction and Hard Negatives Focused Contrastive Learning. 

\textbf{Hard Negative Pairs Construction.} Inspired by the powerful representational abilities of large embedding models, given anchor entity $t$, we employ BGE \cite{bge} to retrieve hard negatives. 

BGE is a text embedding model whose core function is to convert any piece of text (such as a sentence, paragraph, or query) into an embedding vector rich in semantic information. With the rise of Retrieval-Augmented Generation technology \cite{rag,ragsurvery}, the quality of semantic retrieval has become critical to building LLM systems. Due to its outstanding performance in both English and Chinese-English tasks, BGE has quickly emerged as a leading open-source text embedding model worldwide. In this step, we employ BGE to distill hard negative examples with high semantic similarity.

Initially, for each entity in the dataset, we leverage BGE to encode it into an semantic vector and then organize them into a vector DB, denoted as $V$. Next, for each tail entity $t$ in positive triples, we encode it into a query vector and retrieve top-k most semantically similar entities from $V$. Through this process, we get the hard negative set for the anchor entity from the global level. The hard negative set $N_{global}(t)$ is defined as the retrieved top-k most semantically similar entities from $V$ given $t$.

\textbf{Hard Negatives Focused Contrastive Learning.}  For each positive triple, given tail entity $t$, we obtain an accompanying hard negative set $N_{global}(t)$ through the aforementioned distillation steps. We further take the advantage of  $N_{global}(t)$ by integrating it into CL. Formally, we extend $(hr, t)$ to $(hr, t^{+}, t^{-})$, where $hr$ is the query, $t^{+}$ is the corresponding tail entity and $t^{-}$ is the hard negative entity from $N_{global}(t)$. The training loss of global level CL is defined as:

\begin{equation}
    \mathcal{L}_{global}=-log \frac{e^{\phi(hr,t^{+})/\tau}}{\sum_{t^{-} \in N_{global}(t)}e^{\phi(hr,t^{-})/\tau}+e^{\phi(hr,t^{+})/\tau}}
\end{equation}

Overall, on the global scale, we introduce a fine-grained CL method which forces the model to focus on discriminating the anchor entity from extremely similar hard negatives. Subsequent experimental results indicate that leveraging the powerful representational abilities of large embedding models to distill hard negative examples can achieve a substantial improvement compared to prior methods utilizing only the samples in the same batch as negatives.

\subsubsection{Contrastive Learning Training Objective}
Different semantic-aware CL tasks capture distinct aspects of semantic information in KGs. To facilitate knowledge sharing across tasks, we train our model by jointly performing these tasks. The final CL loss is defined as a joint training objective of the local loss and the global loss:

\begin{equation}
    \mathcal{L}_{CL} = w_{1}\mathcal{L}_{local} + w_{2}\mathcal{L}_{global}
\end{equation}
where $w_{i}$ is tunable hyper-parameters for adapting to specific knowledge graph characteristics.

During training phase, we formulate the overall training objective as the sum of the $CL$ loss and the $TL$ loss:

\begin{equation}
    \mathcal{L}_{overall} = \mathcal{L}_{TC}+\mathcal{L}_{CL}
\end{equation}

\subsection{Evaluation Metrics}
To comprehensively evaluate the proposed model, standard metrics balancing prediction accuracy and error distribution were adopted. These metrics are based on four core prediction categories from the confusion matrix, detailed below:
\begin{enumerate}
\item True Positives (TP): Correct positive predictions for actual positive samples.
\item True Negatives (TN): Correct negative predictions for actual negative samples.
\item False Positives (FP): Incorrect positive predictions for actual negative samples.
\item False Negatives (FN): Incorrect negative predictions for actual positive samples.
\end{enumerate}

Based on the above four categories of results, four key comprehensive metrics were employed to evaluate the model’s performance from multiple dimensions:

\textbf{Accuracy} measures overall prediction correctness, defined as the ratio of correctly classified samples to total samples:

\begin{equation*}
    Accuracy = \frac{TP+ TN}{TP + TN + FP + FN}
\end{equation*}

\textbf{Precision} reflects reliability of positive predictions, calculated as the ratio of true positives to all predicted positives:

\begin{equation*}
    Precision = \frac{TP}{TP+FP}
\end{equation*}

\textbf{Recall} measures the model’s ability to identify actual positives, defined as the ratio of true positives to all actual positives:

\begin{equation*}
    Recall = \frac{TP}{TP+FN}
\end{equation*}

\textbf{F1-Score} is harmonic mean of Precision and Recall, balancing false positives and negatives:

\begin{equation*}
    F1~Score = \frac{2 \times Precision \times Recall}{Precision + Recall}
\end{equation*}

\section{Experiment Results}
\label{experiment}
\subsection{Experimental Setup}
Our model is implemented based on Pytorch \cite{pytorch}. The dual-tower encoder is initialized from the pretrained BERT-based-uncased model (English). Using better pretrained language models is expected to improve performance further. Most hyperparameters except learning rate and training epochs are shared across all datasets to avoid dataset-specific tuning. We conduct grid search on learning rate with ranges \{$10^{-5}$, $3 \times 10^{-5}$, $5\times 10^{-5}$\}. Entity descriptions are truncated to a maximum of 128 tokens. Temperature $\tau$ is initialized to 0.05. We use AdamW optimizer \cite{adamw} with cosine learning rate decay \cite{sgdr}. Models are trained with batch size 64 on 2 3090 GPUs.

\subsection{Dataset Description}

We use two datasets for evaluation: FB15k-237, and YAGO3-10. \cite{transE} proposed the FB15k dataset. Later work \cite{fb15k237-1,fb15k237-2} showed that this dataset suffered from test set leakage and released FB15k-237 datasets by removing the inverse relations. The FB15k-237 dataset consists of $\sim$15k entities and 237 relations from Freebase. YAGO3-10 is a subset of YAGO3 \cite{yago3} and is also widely used in many applications. It contains 123,182 entities and 37 relations, with most triples describing attributes related to people, such as citizenship, gender, occupation, and so on.

For textual descriptions, we use the data provided by KG-BERT \cite{kgbert} for YAGO3-10 and FB15k-237 datasets.

\begin{table}[htbp]
\centering
\footnotesize
\setlength{\tabcolsep}{12pt}  
\caption{Statistics of the benchmark datasets.}
\label{tab_dataset}
\renewcommand{\arraystretch}{1.25}
\begin{tabular}{c c c c}
\toprule
\textbf{Dataset}  &  \textbf{\#Train} & \textbf{\#Valid} & \textbf{\#Test} \\
\midrule
  FB15k-237 & 816,345 & 52,605 & 61,398 \\ 
  YAGO3-10  & 3,237,120 & 15,000 & 15,000 \\
\bottomrule
\end{tabular}
\end{table}

\subsection{Data Preprocessing}
The aforementioned datasets are then preprocessed to align with the task formulation of triple classification. For fair competition, we adhere to the setup of \cite{kgeval} and use the same preprocessing method as presented in the original paper: for each dataset, we retain the original positive triples and construct negative triples by randomly replacing the head entity, tail entity and relation, thereby forming a complete triple classification dataset. The resulting dataset statistics are shown in Table~\ref{tab_dataset}.

\subsection{Baseline Approaches}

We compare SASA with four state-of-the-art models: 
\begin{enumerate}
\item TransE \cite{transE}: TransE formulates each relation as a vector translation from the head entity to the tail entity in a low-dimensional space, and then trains these embeddings via a margin-based ranking loss function. 

\item KG-BERT \cite{kgbert}: Leveraging single BERT, KG-BERT takes the textual sequence of triple $(h, r, t)$ as input and then is trained to predict whether the given triple is valid or not.

\item StAR \cite{stAR}: To address the issue of insufficient structured knowledge in text encoders, StAR splits each triple into two components, which are then encoded for contextual representations using a siamese-style text encoder. We modify the model to adapt it to the triple classification task by combining the dual-tower encoder with a vanilla MLP block to assess the correctness of a given triple $(h, r, t)$.

\item Confidence-Based Knowledge Graph Evaluation \cite{kgeval}: The model computes the confidence of a given triple from both local level and global level, addressing the limitations of TransE by leveraging the intrinsic semantic information within KGs.
\end{enumerate}
\subsection{Experimental Findings and Analysis}
\subsubsection{Overall Comparison}

We compare our model with above models on standard benchmarks, including FB15k-237 and YAGO3-10. The overall results are presented in Table~\ref{main_results}.

Based on the accuracy and F1-score, which most accurately depict a model’s total performance, SASA achieves significant improvements over the previous state-of-the-art methods, demonstrating its effectiveness. When compared to Confidence-Based Knowledge Graph Evaluation \cite{kgeval}, SASA achieves an average relative accuracy score improvement of 19.3\% and an average relative F1-score improvement of 16.2\% on FB15k-237, highlighting  text-based models' superiority over embedding-based models. In contrast with text-based state-of-the-art method StAR \cite{stAR}, in which the $(h,r)$ representation and the $t$ representation are simply concatenated and then fed into MLP for TC, SASA performs much better than StAR on both datasets, with a margin of 5.9\% on FB15k-237 and 3.4\% on YAGO3-10, respectively. This result suggests that SASA's outstanding performance does not rely on the increase in model parameters. In fact, the nature of TC is a semantic matching problem between entities and relations rather than a classification problem. But StAR simply models this task as a classification framework. The results show that inappropriate modeling can degrade the model's performance.

The aforementioned improvements emphasize that combining separated attention mechanism with semantic-aware CL can effectively improve the model's performance. Overall, SASA markedly improves upon existing state-of-the-art baselines.

\begin{table}
\centering
\setlength{\tabcolsep}{2pt}
\caption{Experimental results on FB15k-237 and YAGO3-10 datasets for TC. The best performances are highlighted in boldface.}
\label{main_results}
\scriptsize
\renewcommand{\arraystretch}{1.25}
\begin{tabular}{c|cccc|cccc} 
\toprule
\multirow{2}{*}{\textbf{Model}} & \multicolumn{4}{c|}{\textbf{FB15k-237}}   & \multicolumn{4}{c}{\textbf{YAGO3-10}}  \\
\cline{2-9}
& Accuracy   & Precision  & Recall  & F1-score  & Accuracy   & Precision  & Recall  & F1-score  \\ 
\midrule
TransE~\citep{transE}  & 0.699  & 0.712  & 0.678 & 0.695 & 0.718  & 0.722  & 0.726  &  0.724   \\
Confidence-based KGC~\citep{kgeval} & 0.773  &  0.811 & 0.767 & 0.788 & 0.793  & 0.816  & 0.822  & 0.819    \\
KG-BERT~\citep{kgbert}  & 0.860  & 0.835  & 0.876 & 0.855 & 0.928  & 0.896  & 0.885  & 0.891    \\
StAR~\citep{stAR}  & 0.907 & 0.850  & 0.900 & 0.874 & 0.932  &  \textbf{0.936} & 0.895  &  0.915   \\
\midrule
\textbf{SASA (ours)} & \textbf{0.966} & \textbf{0.926} & \textbf{0.974}  & \textbf{0.950}   & \textbf{0.966} & 0.927  & \textbf{0.973} & \textbf{0.949}  \\
\bottomrule
\end{tabular}
\end{table}

\subsubsection{Ablation Study on Various Submodules}

\textbf{Effect of Separated Attention Mechanism.}
We perform experiments to demonstrate the necessity of separated attention mechanism. As illustrated in Table~\ref{ablation_study_results}, the absence of separated attention leads to significant performance degradation on both FB15k-237 and YAGO3-10, which underscores its effectiveness in capturing fine-grained semantic associations between tail entity and head-relation pair.

\textbf{Effect of Local-level CL.}
To validate the effectiveness of our proposed local level CL, we conduct an ablation study. As illustrated in Table~\ref{ablation_study_results}, incorporating local level CL enhances performance across all metrics on both datasets. Compared to $SASA_{SA}$, $SASA_{SA+LL}$ shows a substantial 0.35\% accuracy increase on the FB15k-237 dataset, while the gain on YAGO3-10 is comparable. At the local level, we align the head-relation pair with its dropout-augmented positives, which contributes to endowing the model with a more robust expressive capacity of knowledge representations.

\begin{table}
\centering
\setlength{\tabcolsep}{2pt}
\caption{Ablation study results for FB15k-237 and YAGO3-10 datasets.}
\label{ablation_study_results}
\footnotesize
\renewcommand{\arraystretch}{1.25}
\begin{tabular}{c|cccc|cccc} 
\toprule
\multirow{2}{*}{\textbf{Method}} & \multicolumn{4}{c|}{\textbf{FB15k-237}}   & \multicolumn{4}{c}{\textbf{YAGO3-10}}  \\
\cline{2-9}
& Accuracy   & Precision  & Recall  & F1-score  & Accuracy   & Precision  & Recall  & F1-score  \\ 
\midrule
$SASA_{vanilla}$  & 0.907 & 0.850  & 0.900 & 0.874 & 0.932  &  \textbf{0.936} & 0.895  &  0.915   \\
$SASA_{SA}$  & 0.959  & 0.915  & 0.966 & 0.940 & 0.961  & 0.933  & 0.951  & 0.942    \\
$SASA_{SA+LL}$  & 0.963 & \textbf{0.929}  & 0.961 & 0.945 & 0.965  &  0.922 & \textbf{0.977}  &  0.948   \\
$SASA_{SA+LL+GL}$ & \textbf{0.966} & 0.926 & \textbf{0.974}  & \textbf{0.950}   & \textbf{0.966} & 0.927  & 0.973 & \textbf{0.949}  \\
\bottomrule
\end{tabular}
\end{table}

\textbf{Effect of  Global-level CL.}
In addition to the local level CL, we further enrich the model's training objective with the global level CL. As shown in Table~\ref{ablation_study_results}, on FB15k-237 and YAGO3-10 datasets, removing the global-level CL causes a substantial performance drop, as the model fails to comprehend the subtle semantic differences conveyed by hard negative examples.

The findings above collectively affirm that separated attention, local CL and global CL are all indispensable components for achieving optimal model performance.

\subsubsection{Ablation Study on Negative Sample Quantity}
\begin{figure}
    \centering
    \includegraphics[width=0.68\textwidth]{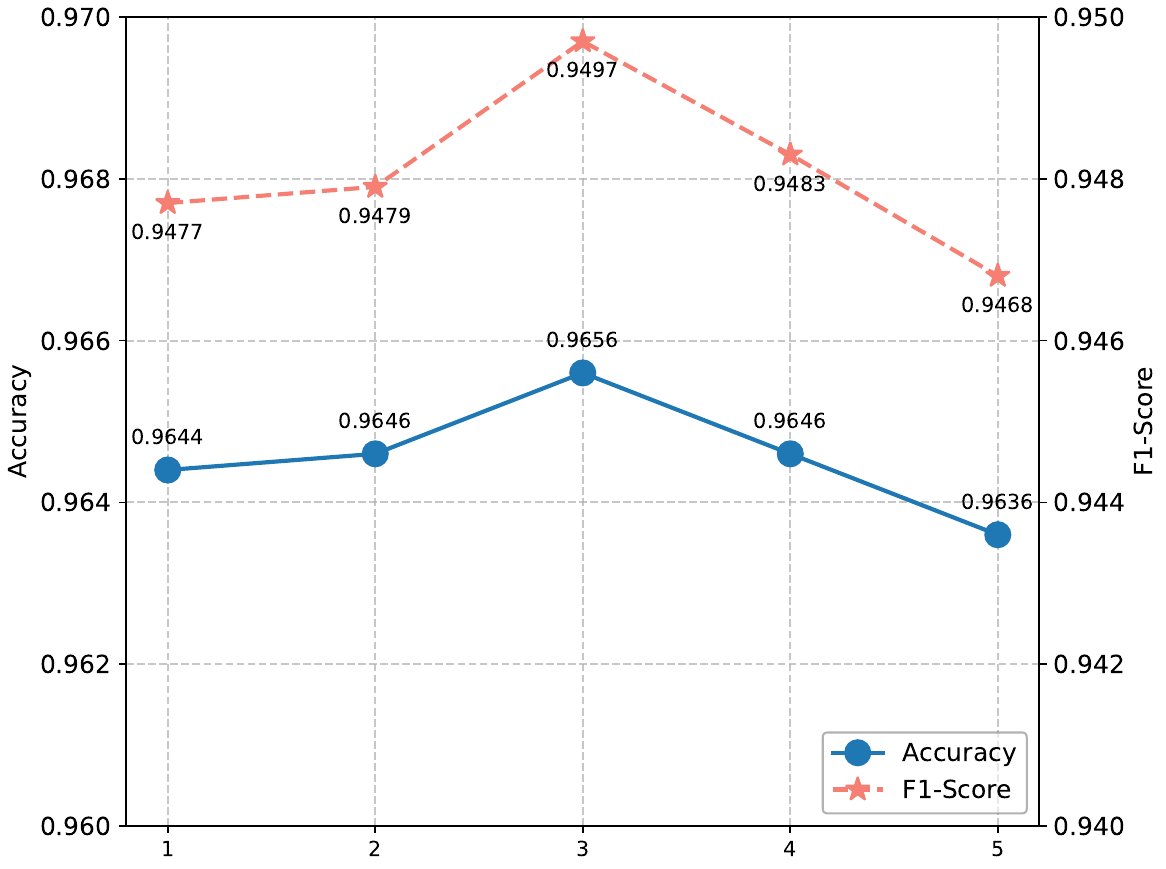}
    \caption{Effect of negative sample quantity.} \label{fig_neg_sample}
\end{figure}

Previous work usually use an in-batch strategy to reuse samples in the same batch as negatives. However, we observe that forcing the model to distinguish between hard negative pairs with extreme semantic similarity can substantially improve its discrimination ability. Thus, we adopt an out-batch strategy which retrieves hard negatives utilizing BGE from data collections for every tail entity. To analyze the contribution of negative samples in the global level CL, we fix the batch size and vary the sampling number of negative samples from 1 to 5 per case to analyze their impact on FB15k-237. There is a clear trend that when sampling number varies from 1 to 3, the performance steadily improves from 96.44\% to 96.56\% in Table~\ref{fig_neg_sample}, which shows that with increasing negative instances within this interval, the model can grasp more fine-grained semantic differences and continually improve its discriminative ability. Nonetheless, when there are more negative samples, we have observed that the model's performance decreased from 96.56\% to 96.36\% and the advantage of adding more negative samples disappeared. We posit that this phenomenon stems from the noise introduced by incorporating additional more negative examples. When employing BGE to retrieve hard negatives, the retrieved samples with lower ranks (i.e., towards the end of the ranking list) are less semantically relevant to the anchor entity, which could introduce noise into the training process. These findings indicate a tradeoff between contextual richness and information noise. Therefore, we set the hard negative number to 3 for the global-level CL in all subsequent experiments.

What's more,  adding more negatives requires more GPU memory and may cause optimization difficulties. We do not experiment with negative number larger than 5.

\begin{table}[htbp]
\centering
\footnotesize
\setlength{\tabcolsep}{8pt}  
\caption{Examples for different categories of relations
on the FB15k-237 dataset.}
\label{relation_example_dataset}
\renewcommand{\arraystretch}{1.25}
\begin{tabular}{c c c c}
\toprule
\textbf{1-1}  &  \textbf{1-M} & \textbf{M-1} & \textbf{M-M} \\
\midrule
  program creator & students & nationality & award winner \\ 
  capital  & actor & gender & major field of study \\
  organizations founded  & film director & place of death & nominated for \\
  spouse  & cause of death & crew member & football roster position \\
  romantic relationship  & member states & place of birth & adjoining relationship \\
\bottomrule
\end{tabular}
\end{table}

\begin{table}[htbp]
\centering
\footnotesize
\setlength{\tabcolsep}{20pt}  
\caption{Proportion of different categories of relations
on the FB15k-237 dataset.}
\label{relation_proportion_dataset}
\renewcommand{\arraystretch}{1.25}
\begin{tabular}{c c c c c}
\toprule
\textbf{Dataset} & \textbf{1-1}  &  \textbf{1-M} & \textbf{M-1} & \textbf{M-M} \\
\midrule
  FB15k-237 & 2.29\% & 7.32\% & 25.31\% & 65.09\% \\ 
\bottomrule
\end{tabular}
\end{table}

\begin{table}[htbp]
\centering
\footnotesize
\setlength{\tabcolsep}{20pt}  
\caption{Accuracy for different kinds of relations on the FB15k-237 dataset with SASA.}
\label{relation_category_results}
\renewcommand{\arraystretch}{1.25}
\begin{tabular}{c c c c c}
\toprule
\textbf{Dataset} & \textbf{1-1}  &  \textbf{1-M} & \textbf{M-1} & \textbf{M-M} \\
\midrule
  FB15k-237 & 0.974 & 0.953 & 0.958 & 0.969 \\ 
\bottomrule
\end{tabular}
\end{table}

\subsubsection{Study of Relation Category}

Following the rules by \cite{transE}, relation mappings in knowledge graphs can be categorized into four groups: one-to-one (1-to-1), one-to-many (1-to-M), many-to-one (M-to-1), and many-to-many (M-to-M). A given relation is 1-to-1 if a head can appear with at most one tail, 1-to-M if a head can appear with many tails, M-to-1 if many heads can appear with the same tail, or M-to-M if multiple heads can appear with multiple tails. We classify the relations into these four classes by computing the averaged number of heads $h$ and tails $t$ appearing in the FB15k-237 train dataset for each relation. If this average number is below 1.5 then the argument is labeled as 1 and M otherwise. For example, a relation having an average of 1.2 head per tail and of 3.2 tails per head was classified as 1-to-M. Table~\ref{relation_proportion_dataset} presents the statistical results in FB15k-237. We find that FB15k-237 test dataset has 2.29\% of 1-to-1 relations, 7.32\% of 1-to-M, 25.31\% of M-to-1, and 65.09\% of M-to-M. Examples are shown in Table~\ref{relation_example_dataset}. 

Now we further analyze the performance of SASA across different relation categories. Based on closer examination of Table~\ref{relation_category_results}, we observe that when it comes to triples with one on the both side, our model exhibits a notable advantage. The reason can be attributed to the fact that 1-to-1 relations typically possess simpler patterns, thus facilitating more efficient model learning. Furthermore, triples with one on the either side present a significantly greater challenge, resulting in a performance degradation of our model. The inherent difficulty in learning complex relation mappings highlights the necessity of developing effective modeling methods. 

\subsubsection{Representation Visualization}
\begin{figure}
    \centering
    \includegraphics[width=0.68\textwidth]{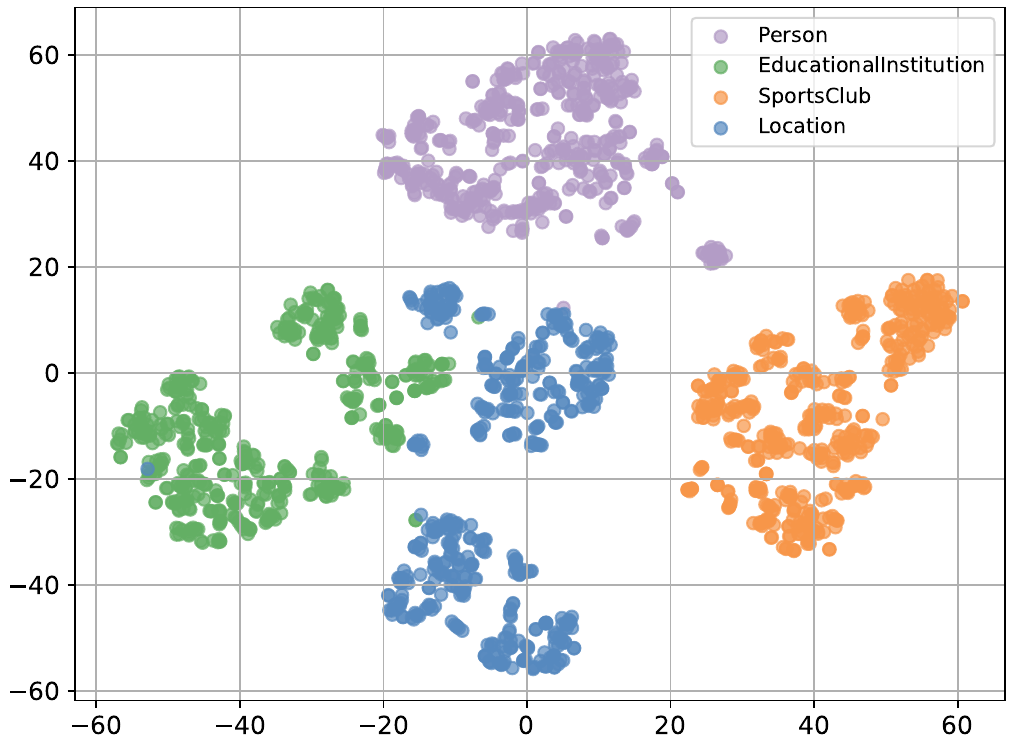}
    \caption{Visualization of entity embeddings.} \label{fig_entity_embedding}
\end{figure}

To provide a qualitative assessment of the representational capabilities of our proposed model, we visualize the entity embeddings, aiming to offer an intuitive examination of whether the model can meaningfully separate entities of different semantic categories in the vector space. Specifically, we randomly sampled 2000 entities across 4 categories in FB15k-237 to ensure a representative and balanced analysis. The entity embeddings were generated with $BERT_{t}$ as detailed in Section~\ref{dual_tower_encoder}. We employ the t-SNE algorithm to project the high-dimensional embeddings onto a 2D plane for visualization. The result, depicted in Figure~\ref{fig_entity_embedding}, reveals that most categories form distinct and well-separated clusters. This clear separation underscores the high quality of the learned embeddings and the model's efficacy in capturing discriminative features. One interesting observation is the partial overlap between the "Educational Institution" and "Location" categories. This overlap reasonably reflects the semantic relatedness between these two concepts in the real world, as an "Educational Institution" can often be considered a type of "Location". This nuance demonstrates our model's ability to encode subtle semantic relations.

\section{Conclusion and Future works}
\label{conclusion}
This study proposes a novel framework for the triple classification task on FB15k-237 and YAGO3-10 datasets. Our proposed approach, named SASA, employs separated attention mechanism for a more effective representation fusion. Furthermore, the integration of semantic-aware contrastive learning including both local level and global level modeling further enhances the model's discrimination ability, resulting in high accuracy, precision, recall, and F1-score on both datasets. These results demonstrate that the combination of of separated attention and semantic aware contrastive learning offers a robust solution for triple classification.

In the future, we plan to investigate how to explicitly integrate graph structural information into PLMs through CL, so as to fully leverage the prior linguistic information of pretrained models and graph structural information.
\bibliographystyle{elsarticle-num} 
\bibliography{cite-num}






\end{document}